\documentclass[a4paper,conference]{IEEEtran}
\usepackage{times}

\usepackage{soul}
\usepackage{url}
\usepackage[hidelinks]{hyperref}
\usepackage[utf8]{inputenc}
\usepackage[small]{caption}
\usepackage{graphicx}
\usepackage{amsmath}
\usepackage{booktabs}
\usepackage{graphicx}
\usepackage{amsmath,amssymb} 
\usepackage{color}
\usepackage{todonotes}

\usepackage{algorithm}
\usepackage{algpseudocode}  
\usepackage{multirow}
\usepackage{booktabs}
\usepackage{subfigure}
\usepackage{hyperref}
\usepackage{color,soul}
\usepackage{amsmath}
\usepackage{hhline}
\usepackage{multicol}
\usepackage[normalem]{ulem}

\ifCLASSINFOpdf
\else
\fi
\usepackage{hyperref}

\begin{document}
%
\title{Domain-specific Learning of Multi-scale Facial Dynamics for Apparent Personality Traits Prediction}




%
\author{\IEEEauthorblockN{Fang Li\IEEEauthorrefmark{1}}
\IEEEauthorblockA{\IEEEauthorrefmark{1}School of Communication and Information Engineering\\
Shanghai Technical Institute of Electronics and Information,
Shanghai, China\\ Email: lf1266@163.com}
}



\maketitle


\begin{abstract}

Human personality decides various aspects of their daily life and working behaviors. Since personality traits are relatively stable over time and unique for each subject, previous approaches frequently infer personality from a single frame or short-term behaviors. Moreover, most of them failed to specifically extract person-specific and unique cues for personality recognition. In this paper, we propose a novel video-based automatic personality traits recognition approach which consists of: (1) a \textbf{domain-specific facial behavior modelling} module that extracts personality-related multi-scale short-term human facial behavior features; (2) a \textbf{long-term behavior modelling} module that summarizes all short-term features of a video as a long-term/video-level personality representation and (3) a \textbf{multi-task personality traits prediction module} that models underlying relationship among all traits and jointly predict them based on the video-level personality representation. We conducted the experiments on ChaLearn First Impression dataset, and our approach achieved comparable results to the state-of-the-art. Importantly, we show that all three proposed modules brought important benefits for personality recognition.

\end{abstract}

\IEEEpeerreviewmaketitle





\section{Introduction}
\label{sec:intro}








\noindent Personality is defined as behavioral and emotional characteristics that partially defines a person's identity and distinguishes the person from others \cite{hogan1997handbook}. Accurately recognizing human personality can benefit various real-world applications, such as employment \cite{zimmerman2008understanding,eddine2017personality}, mental health diagnosis \cite{vukasovic2015heritability,jaiswal2019automatic}, shopping preference \cite{tsao2010exploring,heinstrom2000impact} as well as understanding human cognition and emotional processes \cite{komulainen2014effect,song2021learning}. In the past decades, psychologists have proposed many traits-based personality models to evaluate some specific aspects of human personality, which are stable over time for the each person but differ across individuals \cite{kassin2003essentials}. Among these traits-based models, the 'Five-Factor' model (also called the 'Big-Five' model) \cite{mccrae1987validation} has been widely used in recent years, which reflects five aspects of human personality, i.e., Extraversion (EX), Agreeableness (AG), Openness (OP), Conscientiousness (CO), and Neuroticism (NE).

However, the majority of traditional personality traits assessments are based on an individual's own verbal report (e.g., questionnaires/inventories \cite{john1991big}), which is subjective and not reliable as the subject may provide incorrect answers to hide some crucial information on some occasions (e.g., interviews). Thus, objective solutions that can accurately reflect human personality from their non-verbal behaviors is a potential alteration. Considering that many psychological studies suggested that personality traits can be reflected by human facial behaviors \cite{keltner1996facial,shevlin2003can,knapp2013nonverbal}, face video-based automatic personality analysis has drawn a lot of attention in recent years.

A popular and standard video-based solution predicts personality traits from individuals' appearance or short-term non-verbal visual behaviors. These methods \cite{ventura2017interpreting,guccluturk2016deep,wei2018deep,celiktutan2017automatic,principi2019effect,tellamekala2022dimensional} individually feed each pre-processed visual image or short segment of the video into machine learning (ML) models to predict the corresponding subject's personality traits. In other words, such methods usually use the personality annotations that represent the personality traits reflected by the whole video as the label for each image/short segment, and attempt to train ML models that can predict personality from an image or a short segment. This is problematic because a single image/short segment of different videos (with different personality labels) may carry similar facial behavior information, resulting in that same input patterns paired with different labels during the training, which may not allow the ML model to learn a good hypothesis \cite{song2020spectral,song2018human}. Instead, this training strategy may lead the ML model focus on learning the identity information rather than facial behaviors as the identity is an invariant attribute of the video while short-term facial behaviors usually vary a lot in a video. Besides, all these methods failed to use the long-term behaviors of subjects, which may contain crucial clues for personality recognition. In this paper, we define the long-term behaviors as the video-level behaviors.

While personality traits is defined as human attributes that stable over time, some studies attempted to model long-term visual information for personality recognition. Instead of inferring personality using features only extracted from a single frame/short video segment, these methods proposed to construct a video-level descriptor to represent long-term facial behavior information. While a simple solution \cite{fang2016personality,eddine2017personality} to achieve this is to compute statistics of frame-level descriptors of the entire video, many recent deep learning-based approaches \cite{li2020cr,bekhouche2017personality,zhang2019persemon,beyan2019personality} proposed to deep learn personality-related video-level descriptors from the pre-selected key frames of the video. However, during the key frames selection, these deep learning methods discard a large number of frames, i.e., crucial personality-related behavior details may be discarded. To encode all frames of video into a video-level representation, Song et al. \cite{song2021self,song2021learning} and Shao et al. \cite{shao2021personality} proposed to use person-specific CNN weights as the personality descriptor. However, the main drawback of such methods is that they need self-supervised learn a person-specific CNN or a set of person-specific layers for each test video and thus are not efficient and not appropriate for the fast personality assessment.

\begin{figure*}
  \begin{center}
  \includegraphics[width=16.8cm]{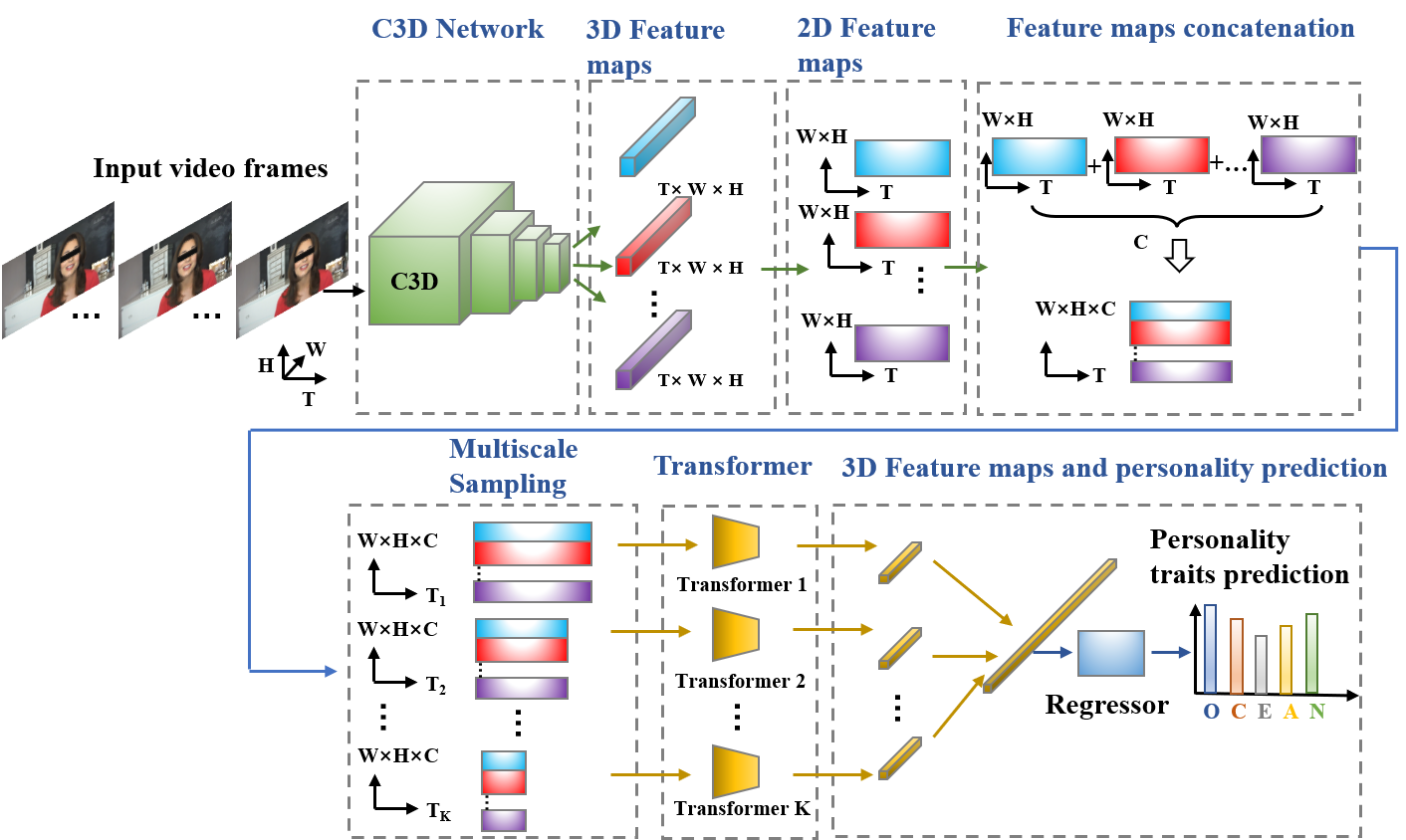}
  \caption{The full pipeline of the proposed personality recognition approach.}
  \label{fig:c3d}
  \end{center}
\end{figure*}

Although long-term facial behaviors are more reliable sources for personality recognition, short-term behaviors may also contain informative clues (this is evidenced by that some short-term automatic modeling methods \cite{wei2018deep,celiktutan2017automatic} also achieved good performance in recognizing apparent personality traits). Motivated by this, this paper propose to deep learn personality-related clues from both short-term and long-term facial behaviors for automatic apparent personality prediction. Firstly, we propose a C3D-Transformer network as the backbone to extract multi-scale spatio-temporal personality-related facial clues from every short segments of the video, where domain-specific learning \cite{bousmalis2016domain} is extended to allow the proposed C3D-Transformer only learning personality-related features from the input. Since the lengths of videos are usually arbitrary in the real-world applications, we utilize the spectral algorithm \cite{song2020spectral,song2018human} to encode all short-term descriptors as a long-term video-level facial behavior representation, which again encode multi-scale video-level behavioral temporal information. Finally, we train a personality traits prediction model that jointly takes each short-term descriptor as well as its corresponding video-level descriptor as the input, where domain-specific learning is again employed to further remove the noises from the input features. The pipeline of the proposed approach is depicted in Fig. \ref{fig:c3d}. The novelty and contributions of the paper are listed as follows: 
\begin{itemize}

    \item We propose a novel C3D-Transformer network that can deep learn personality-related multi-scale short-term spatio-temporal facial clues. To the best of our knowledge, this is very first work that combines C3D and transformer for learning spatial-temporal personality-related facial behavior features. The experimental results shows that the proposed C3D-Transformer outperforms the C3D/ResNet networks that has been frequently used for short-term personality visual feature extraction.
    
    \item We propose a novel domain-specific learning strategy to specifically remove the personality unrelated noises (especially the identity information) from both short-term and long-term facial behavior features.

    \item We empirically evaluate the proposed approach on the ChaLearn dataset and achieved comparable performance to the state-of-the-art.
    
\end{itemize}

\section{Related Work}

\subsection{Psychological relationship between personality and facial behaviors}


\noindent A large number of studies show that humans are capable of interpreting personality from previous non-seen static face [1-7] or spatio-temporal facial behaviors. For example, the symmetry of the face signifies the extraversion of a person [5]. The indicator of facial width to height ratio (fWHR), is shown with correlations of various personality traits, such as achievement striving [6], dominance [7], aggressiveness [8-11], and risk-taking [12]. The stability of the FWHR indicator, regardless of the hair, means that static facial traits can convey enough personality characteristics [13]. Meanwhile, other facial attributes, such as wrinkles or smile lines, are also shown to be informative for personalities. Since these facial muscle formations are developed by personality relevant habitual emotional facial expressions [3]. The facial expression of smile is often used as cue to judge a person’s extraversion, neuroticism, and agreeableness while cheerful facial expressions are often perceived to influence the level of extraversion in Big Five. In [19], researchers found that gaze behaviors (either in direction of gaze, blink rates or eye contact) can reflect the personality characteristics, and preference of direct gaze avoidance and subjective averted gaze, are often perceived with higher neuroticism scores. In addition, eople with more eye contact, vocal variety, gesticulation, and facial expressions, are often interpreted with high score of extraversions [22]. In terms of temporal behaviors, people with fixated longer duration of fearful eye-region faces, are perceived with high score of neuroticisms, compared to those with shorter duration of that [21].


\subsection{Video-based automatic personality analysis approaches}

\noindent In this section, we explicitly review the existing video-based automatic apparent personality traits recognition methods. As discussed in Sec. \ref{sec:intro}, these methods can be roughly categorized into two types based on their temporal encoding strategies: methods that predict personality from a frame or a short video segment and methods that predict personality from the whole video.

For the frame/short segment-level methods, they usually predict personality traits from each frame or short segment of the video and then fuse all frame/segment-level predictions as the personality prediction of the video (the subject). In \cite{wei2018deep}, each video is represented by about 100 images, each of which is then processed by their DAN (Descriptor Aggregation Network) model, individually, to generate frame-level predictions. Then, they average all frame-level predictions to make the video-level personality prediction. Ventura et al. \cite{ventura2017interpreting} added a CAM module to the DAN network for visualizing the most important face part during personality recognition. Similar to \cite{wei2018deep}, this method also extracts frame-level features and then averages the frame-level predictions as the video-level personality prediction. In \cite{guccluturk2016deep}, a random frame in a video is selected as the video representation, which is then fed into a Deep Residual Network. Celiktutan et al. \cite{celiktutan2017automatic} proposed to extract several low-level hand-crafted features from each frame, which are then combined to infer personality traits. Besides, some other methods proposed to extract image-level facial landmarks \cite{vernon2014modeling} or human mid-level attributes \cite{joo2015automated} to understand the human personality traits. However, all the aforementioned methods only attempted to infer personality from either a single frame or a thin slice (less than 1 second), which are not reliable as these would cause a classic machine learning issue, the same input pattern has multiple labels, resulting in poor generalization capability. In addition, a single frame or a short video segment may be too short to carry enough and reliable information for recognizing an individual's personality.

Therefore, many recent studies have been devoted to infer personality from long-term information, i.e., the whole video. Aran et al. \cite{aran2013one} extract hand-crafted per-frame/segment features from a one-minute video for each person, such as visual activity features, motion template-based features, and visual focus of attention features, to predict personality traits. Fang et al. \cite{fang2016personality} use normalized hand-crafted interactive features (e.g., intra-personal features, dyadic features, and one-vs-all features) to represent the video-level personality information, which are processed by Support Vector Machine (SVM) and Ridge Regression for personality recognition. In \cite{eddine2017personality}, a mean vector of Pyramid multi-level Binarized Statistical Image Features (PML-BSIF) and a PML Local phase Quantization (PML-LPQ) method are proposed to summarize all frames' information of a video. Both video-level representations are then fed to a machine learning system to evaluate job candidates' personality traits. 

Besides, the deep learning-based approach attracted more attention in recent years. One of the most popular solutions for learning video-level personality descriptor is to downsample a whole video into a set of representative frames and then summarize these frames or their features into a unified representation. For example, in \cite{li2020cr}, each video is down-sampled to 32 frames for both global scene and local face region, respectively. Then, these frames are jointly fed to a ResNet-based framework to estimate the five traits at the video level. Zhang et al. \cite{zhang2019persemon} also select a face frame from each thin slice, and then concatenate all selected frames as the video-level representation. In this work, they employed a consensus strategy to jointly process all selected frames for video-level personality prediction. Instead of discarding a large number of frames as \cite{li2020cr,zhang2019persemon}, Song et al. \cite{song2021self} proposed a domain adaptation method that encodes all frames information of a video into a set of intermediate CNN layers that contains person-specific dynamics of the corresponding subject. Then, they use the weights and biases of the learned intermediate CNN layers as the person-specific representation for personality recognition. Additionally, they also propose to search a person-specific CNN architecture for each individual, representing the cognitive process of the target individual \cite{song2021learning,shao2021personality}. Similarly, each person-specific CNN architecture is then parametrized as a video-level graph representation which is fed to Graph Neural Networks for personality recognition.

\section{The proposed approach}

In this section, we present the proposed approach that deep learns both multi-scale short-term and long-term personality-related facial dynamics for automatic personality traits recognition. Specifically, we first propose a C3D-Transformer network as the backbone to extract multi-scale short-term facial behaviors for each frame of the given video (Sec. \ref{subsec: C3D-trans}), where the domain-specific learning is proposed here to enforce the well-trained network to only learn personality-related information rather than noises such as identities (Sec. \ref{subsec: DS}). To encode a video-level representation, we further propose to summarize all deep-learned short-term facial features of the given video into a spectral representation that encodes multi-scale facial temporal information (Sec. \ref{subsec: long-term}). Finally, we feed the produced spectral representation into an ANN model for personality traits prediction, where the domain-specific learning is again employed to remove the personality-unrelated information from the input video-level spectral representation (Sec. \ref{subsec: personality model}).

In comparison to previous approaches, the main advantages of our method are: 1) Unlike \cite{ventura2017interpreting,guccluturk2016deep,wei2018deep,celiktutan2017automatic,principi2019effect} failed to consider multi-scale facial temporal information, our method proposes C3D-Transformer network to model multi-scale short-term temporal dependencies between frames, which is crucial for representing facial behaviors; 2) we present the first work that specifically separates the learned short-term facial behaviour features and video-level representations into two parts: personality-specific features and unrelated noises; 3) Different from \cite{li2020cr,zhang2019persemon} that remove a large number of frames, we propose to encode all short-term facial behavior features extracted from all available frames into a video-level representation which retains multi-scale video-level facial dynamics.

\subsection{Short-term facial dynamics modelling}

\subsubsection{C3D-Transformer network architecture}
\label{subsec: C3D-trans}

While a facial image only contain a static facial display and the corresponding person's identity, the short-term facial behavior carried by an image sequence can provide more discriminative spatio-temporal behavioral information for personality recognition. While some previous methods \cite{subramaniam2016bi} employed short-term facial behaviors for personality recognition, they used RNNs/LSTMs to model facial dynamics from low-dimensional latent features, which may ignore crucial dynamic information during the feature extraction process.

To avoid the aforementioned problems, this paper proposes a C3D-transformer network to model multi-scale short-term facial dynamics from the original video segments. As shown in Fig. \ref{fig:c3d}, the C3D-transformer consists of a 3D convolution block which ensure the facial behavioral information within multiple frames can be directly extracted without using LSTM/RNNS. To be specific, the employed C3D block consists of three convolution blocks, where each contains a 3D convolution layer, a 3D batch normalization layer and a ReLU activation function to simulate non-linear projections. For a video segment, the output of the C3D block is a set of 3D feature maps, which are then aggregated as a single multi-channel time-series feature map. Let us assume that the produced 3D feature maps of size $C \times T \times W \times H$, where $C$ is the number of feature maps while $T$, $W$, $H$ are the temporal scale, width and height of each feature map. The size of the aggregated multi-channel time-series feature map is $CWH \times T$, i.e., the time series has $CWH$ channels and $T$ frames. After that, the transformer module firstly down-samples the produced multi-channel time-series feature map along the temporal dimension. Consequently, multiple multi-channel time-series feature maps with temporal scales of $T_1, T_2, \cdots, T_K$ can be produced, each of which are then fed to a fully connected layer to generate a 1D representation, respectively. As a result, these 1D representations contain multi-scale temporal information of the input video segment. Finally, multiple multi-head transformers are introduced to individually process each of these 1D representations, and the outputs of all transformers are concatenated as the segment-level representation. In comparison to previous approaches which failed to consider multi-scale facial temporal information, the proposed C3D-Transformer can model multi-scale short-term temporal dependencies between frames, which is crucial for representing facial behaviors;

\subsubsection{Domain-specific training strategy}
\label{subsec: DS}

Although the proposed C3D-transformer can already deep learn personality-related features from the input data, as discussed in Sec. \ref{sec:intro}, pairing video-level labels with short segments may lead the network focus on learning some personality-unrelated noises, which are invariant attributes of the person in the video, e.g., identity, and ignoring their spatio-temporal behaviors. To solve this problem, this section proposes a domain-specific learning strategy to remove such noises. 

\begin{figure*}
  \begin{center}
  \includegraphics[width=16.8cm]{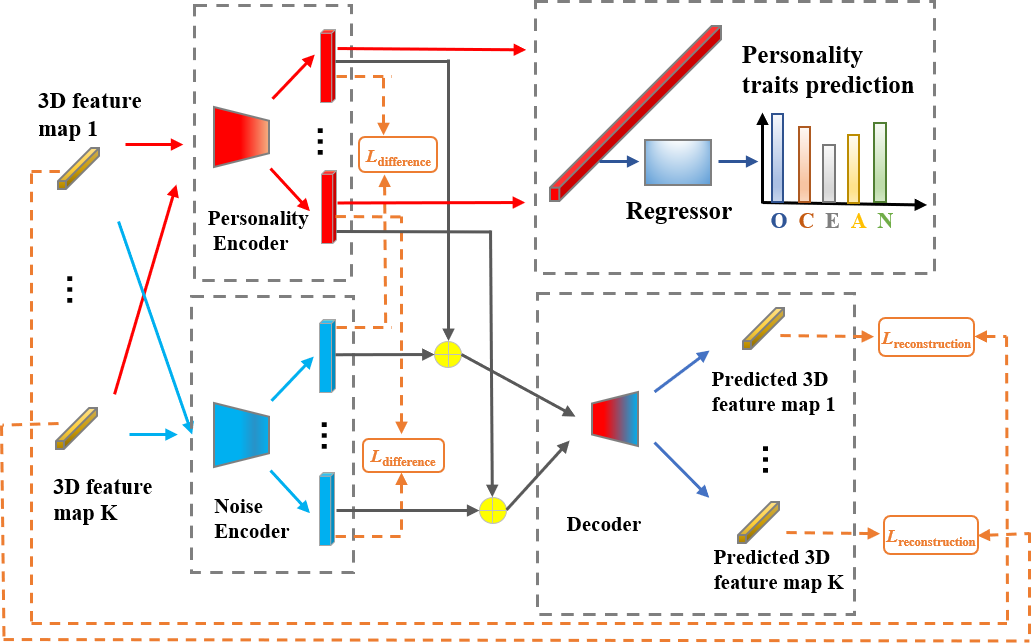}
  \caption{Illustration of the proposed domain-specific learning (DS) module for each trait. It should be noted that our full pipeline consists of five DS modules, where each module learns a representation for predicting a individual trait.}
  \label{fig:domain-sepration}
  \end{center}
\end{figure*}

As shown in Fig. \ref{fig:domain-sepration}, during the training, the representations produced by all transformers are individually fed to five pairs of encoders, where each pair contains a personality encoder that corresponds to a specific personality trait and a noise encoder to extract noises that un-relate to the personality traits. In other words, the assumption is that the extracted feature at each temporal scale (from each transformer) is made up of two parts: personality trait-related feature and un-related noises. In particular, all encoders consist of two fully connected layers with a ReLU and a dropout attached. To train them as well as the C3D-Transformer, several loss functions are introduced. Firstly, we feed each output of the personality encoder to a classifier in order to jointly predict the corresponding five personality traits, which is supervised by the MSE loss:
\begin{equation}
\label{eq:mse}
\text{Loss}_{1} = \sum_{i=1}^5 \left({pd}_{i} - label_{i} \right)^{2}
\end{equation}
where ${pd}_{i}$ is the $i_{th}$ personality traits prediction and $label_{i}$ is the corresponding label of the video segment. This loss would enforce all personality encoders to specifically learn personality-related information from the output of the C3D-Transformer. Meanwhile, each personality feature extracted from the personality encoder should be dissimilar to the corresponding unrelated noises, which can be denoted as:
\begin{equation}
\label{eq:dis}
\text{Loss}_{2} =  \sum_{i=1}^5 \left\|({fea}^{P-Enc(i)})^{\top} {fea}^{N-Enc(i)} \right\|_\text{Frob}^{2}
\end{equation}
where ${fea}^{P-Enc(i)}$ is the personality feature extracted from the $i_{th}$ personality encoder, and ${fea}^{N-Enc(i)}$ is the noise extracted from the noise encoder. $\|\cdot\|^{2}_F$ is the square Frobenius norm.

To further ensure that the personality-related features and un-related noises are separated by each pair of encoders, we additionally introduce five decoders. Each encoder also consists of two fully connected layers, and each fully connected layer is followed by a ReLU and a dropout layer. A decoder takes features generated from the corresponding personality and noise encoders, targeting to reconstruct the original temporal feature generated by the corresponding transformer. This process can be formulated as:
\begin{equation}
\text{Loss}_{3} = \frac{ \sum_{d=1}^{D} \left(fea_i^d(\text{Dec})-fea_i^d(\text{Tran})\right)^{2}}{I \times D} 
\end{equation}
where $fea_i^d(\text{Dec})$ and $fea_i^d(\text{Tran})$ are the $k_{th}$ element of the reconstructed feature corresponding to $i_{th}$ trait and the $k_{th}$ element of the $i_{th}$ transformer's output. In comparison to \cite{bousmalis2016domain}, we disregard the similarity loss that forces personality-related features generated by each input samples to be the same. This is because we do not want that the input samples at each batch are required to have the same personality traits (which is impossible). As a result, the final loss function to train the C3D-Transformer and the personality-related encoders is defined as: 
\begin{equation}
\label{eq:rec}
\text{Loss}_{\text{overall}} = \alpha \text{Loss}_{1} + \beta \text{Loss}_{2} + \gamma \text{Loss}_{3}
\end{equation}
where $\alpha$, $\beta$ and $\gamma$ are corresponding weights for each loss.

\subsection{Long-term facial dynamics modelling}
\label{subsec: long-term}


\noindent After obtaining frame-level or short video segment-level predictions, most previous studies compute the video-level prediction by averaging all frame-level predictions or combining short segment-level predictions. However, such simple strategies failed to consider the long-term behavioral temporal information. In this paper, we first employs the spectral heatmaps proposed by \cite{song2018human,song2020spectral} to re-summarize multi-scale behavioral temporal information from all segment-level features of the given video, to construct a multi-scale video-level behavioral representation. 

Supposing that for an given video with $N$ short segments, the proposed short-term modelling module extracts $N$ features with $D$ dimensions, the video-level spectral heatmaps is constructed as follows.
\begin{itemize}
    
    \item \textbf{Step 1:} Arranging all segment-level features as a multi-channel time-series signal with $D$ channels and $N$ time stamps.
    
    \item \textbf{Step 2:} Converting each channel of time-series as a spectral signal with $N$ dimensions using Discrete Fourier Transform, resulting in a $D \times N$ amplitude map and a $D \times N$ phase map.
    
    \item \textbf{Step 3:} As discussed in \cite{song2020spectral}, most behavior information are retained by low frequency components. Thus, we then select only top-M lowest frequencies from both amplitude map and phase map. Consequently, a $D \times M$ amplitude map and a $D \times M$ phase map can be generated. Since both amplitude and phase maps are symmetric, the $M < \frac{N}{2}$ is set here.

\end{itemize}
In short, the produced amplitude and phase maps contain $M$ video-level frequencies, where the values of each frequency components indicate behavioral information of a unique temporal scale. This is to say, these two spectral maps are video-level representations that encode $M$ scales behavioral dynamics of the given video. In this paper, we concatenate two the amplitude and phase maps as a two-channel heatmap.

\subsection{Multi-task personality traits prediction module}
\label{subsec: personality model}

\begin{figure*}
  \begin{center}
  \includegraphics[width=16.8cm]{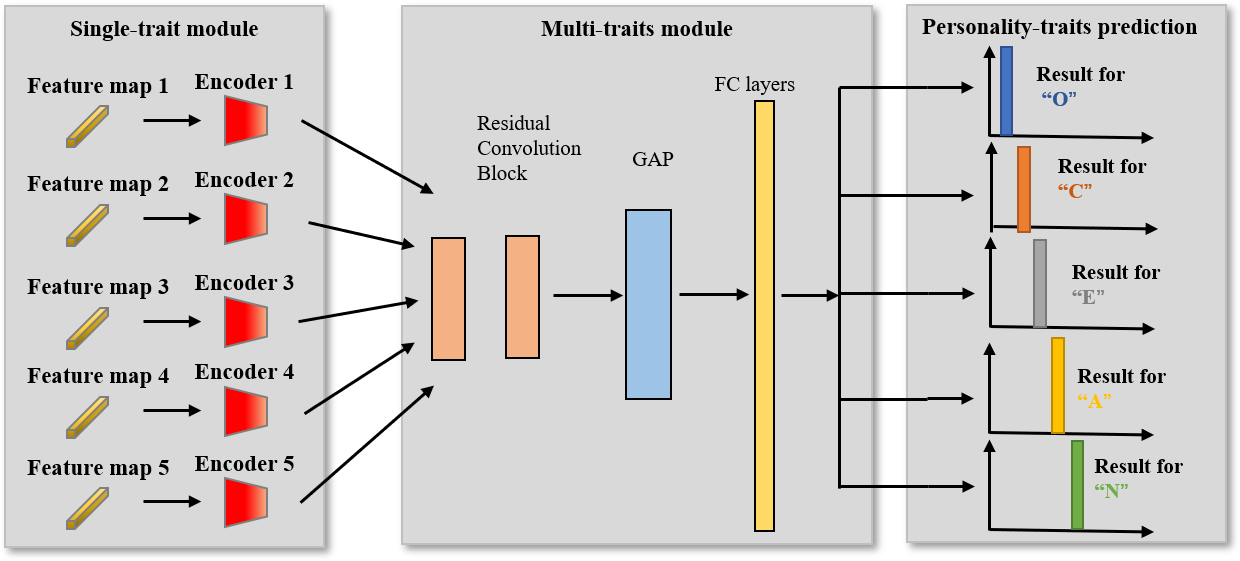}
  \caption{Multi-task personality traits prediction model.}
  \label{fig:multi-task}
  \end{center}
\end{figure*}

\noindent In this section, we present our personality traits prediction model for jointly predicting five personality traits. As shown in Fig. \ref{fig:multi-task}, the model consists of two modules: single-trait module and multi-traits module. The single-trait module contains five branches, each of which contains three 1D convolution blocks (a 1D convolution layer, a dropout layer and a ReLU) to generate the trait-specific feature from the input heatmaps independently. Then, the multi-traits module that contains two 1D residual convolution blocks, a global average pooling layer and three fully connected layers are proposed to combine five sets of trait-specific feature maps and model the underlying relationship among them, in order to further enhance the prediction performance.

During the training, each branch of the single-trait module is connected with two fully connected layers and aims to output predictions of the corresponding trait. This would provide intermediate supervision for each branch, enforcing each of them focusing on learning features that only relates to the target personality trait, which is denoted as:
\begin{equation}
\text{Loss}_{\text{single}} = (pd_i^{single} - label_i)^2
\end{equation}
Meanwhile, we also attach another Mean Square Error loss function to evaluate the five traits' predictions generated by the multi-trait module. This loss is defined as 
\begin{equation}
\text{Loss}_{\text{multi}} = \sum_{i=1}^5 (pd_i^{multi} - label_i)^2
\end{equation}
As a result, the utilized loss functions can not only allow each single-trait branch to focus on modelling the trait-related information but also the multi-trait module to further learn the relationship among these extracted trait-specific features, providing more valuable clues for personality traits prediction.

\section{Experiments}

\noindent In this section, we first list and explain the database used for evaluation in Sec. \ref{subsec:database}. Then, the evaluation metrics are detailed in Sec. \ref{subsec:Metrics} and we present all model settings and training details in Sec. \ref{subsec:Implementation}. After that, the experimental results including comparison to other methods as well as ablation studies are listed and discussed in Sec. \ref{subsec:results}.

\subsection{Database}
\label{subsec:database}

\noindent Apparent personality estimation experiments were conducted on the ChaLearn \cite{ponce2016chalearn} database. This database records $10,000$ talking-to-the-camera videos from $2,764$ YouTube users, where each video is $15$ seconds with fps of $30$. It has been divided to three subsets: a training set (contains $6,000$ videos), a validation set (contains $2,000$ videos) and a test set (contains $2,000$ videos). Each video is labeled with six dimensionas e.g., Big-Five personality traits and the interview' dimension, using Amazon Mechanical Turk by several human annotators, and the ground-truth of each dimension ranges from 0 to 1.

\subsection{Metrics}
\label{subsec:Metrics}
\noindent To compare our approach with other methods on the ChaLearn dataset, the widely-used \textbf{mean accuracy} measurement $\mbox{ACC}$ \cite{ponce2016chalearn} is employed:
\begin{equation}
\mbox{ACC} = 1 - \frac{1}{N}\sum_{i = 1}^{N}|p_i-g_i|,
\end{equation}
where $g_i$ and $p_i$ are the labels and predictions, respectively and $N$ is the number of videos.

\subsection{Implementation details}
\label{subsec:Implementation}

\noindent \textbf{Pre-processing:} To obtain the face region from each frame, we used OpenFace 2.0 \cite{baltrusaitis2018openface} to process all frames of each video, and used the aligned face images as the input to our C3D-Transformer network.

\textbf{Model settings:} In this paper, the multi-scale encoding module contains two down-sampling rates, i.e., choosing a frame for every $2$ frames and $5$ frames. As a result, we used three transformers that take the original feature map and two down-sampled feature maps, respectively. Each transformer network consists of two self-attention layers and four fully connected layers with six heads, where the last fully connected layer output a $64$ dimension vector as the segment-level personality representation. We set the spectral heatmaps that are used for encoding long-term to keep top-32 selected frequencies for each channel, result in a $64 \times 32$ amplitude map and a $64 \times 32$ phase map for each video. The probability of dropout layers in the personality prediction model are all set to $0.5$.

\textbf{Training details:}  We first train the C3D network by pairing the video-level label with the each short segment using MSE loss function and Adam optimizer with learning rate of $0.005$, and then re-train the entire C3D-Transformer in an end-to-end manner, where we re-used Adam optimizer with learning rate of $0.001$. The batch size for training C3D and C3D-Transformer are both set to $3$, and at each batch, there are $30$ frames. Then, the personality prediction model (the MLP) is trained using SGD optimizer with learning rate of $0.0005$ and batch size is set to $64$ spectral heatmaps. In this paper, we used the pre-defined $6000$ videos for training, $2000$ videos for validation and $2000$ videos for testing. All experiments and code were implemented based on PyTorch.

\subsection{Results}
\label{subsec:results}

\noindent In this section, we first compare our best system with the existing approaches in Sec. \ref{subsec:sota}, showing the competitive performance of the proposed approach. Then, we report the results achieved by a set of ablation studies in Sec. \ref{subsec:ablation}, which provide the explicit evaluation for the influence of the proposed C3D-Transformer, domain-specific learning, video-level spectral encoding and multi-task learning modules on the recognition performance.

\subsubsection{Comparison with existing methods}
\label{subsec:sota}

\noindent We compare our best system which consist of three modules: C3D-Transformer learned with the domain-specific training strategy, spectral-based video-level feature encoding, and the multi-task regressor, to other recently proposed approaches in Table \ref{tb:chalearn}. In addition, we also report the results achieved by the same models trained without domain-specific strategy. It is clear that our best system achieved comparable average result to state-of-the-arts methods, i.e., our average performance (0.9163) is very close to the best approach (0.917) \cite{zhang2019persemon} and the second best approach (0.9168) \cite{song2021self}. More specifically, our approach has promising performances in recognizing the Agreeableness and Openness traits, which achieved the best recognition result for Openness trait and the second best result for Agreeableness trait. In addition, the proposed approach also beat most listed approaches in recognizing Conscientiousness and Neuroticism traits.

These results indicate that the proposed framework can provide relative reliable predictions for most personality traits (Our approach is the state-of-the-art face-based system to recognize the Openness trait) only from individuals' non-verbal facial behaviors. Although two approaches \cite{zhang2019persemon,song2021self} achieved slightly better average performance than our approach. One of them requires extra the use of emotion data and annotations \cite{zhang2019persemon} while the other suffers from very long person-specific layer training duration \cite{song2021self}. In contrast, the proposed approach neither requires extra data and annotation nor needs long training time. In other words, the proposed approach not only can provide similar recognition performance as \cite{zhang2019persemon,song2021self} but also much easier to be implemented.

\setlength{\tabcolsep}{2pt}
\begin{table*}[t!]
	\begin{center}
\resizebox{1\linewidth}{!} {
		\begin{tabular}{|l| c c c c c c|}
			\toprule
Methods & Extra & Agree & Consc & Neuro & Open & Avg.   \\
\hline \hline

Baseline \cite{escalante2020modeling} & 0.9019 & 0.9059 & 0.9073 & 0.8997 & 0.9045 & 0.9039  \\

PML \cite{bekhouche2017personality}  & 0.9155 & 0.9103 & 0.9137&  0.9082& 0.9100 &0.9115  \\

NJU-LAMDA  \cite{wei2018deep} $^{*}$ & 0.9112 & 0.9135 & 0.9128 & 0.9098 & 0.9105 & 0.9116  \\

DCC \cite{guccluturk2016deep} $^{*}$ & 0.9088 & 0.9097 & 0.9109 & 0.9085 & 0.9092 &0.9109  \\

PAL \cite{song2021self} & 0.9183 & \textbf{0.9262} & 0.9082  & 0.9133 & 0.9180 & 0.9168  \\

PerEmoN \cite{zhang2019persemon} & \textbf{0.920} & 0.914 & [0.921] & [0.914] & 0.915 & [0.917]  \\
\hline

Ours (Non-DS)   & 0.9135 & 0.9116 & 0.9082 & 0.9107 & 0.9159 & 0.9120 \\

Ours (DS)   & 0.9138 & [0.9190] & 0.9166 & 0.9123 & \textbf{0.9198} & 0.9163 \\

			\bottomrule
		\end{tabular}
        }
	\end{center}
	\caption{ACC results achieved by the proposed approaches and existing approaches on ChaLearn First Impression Dataset, where the bold numbers denote the best result and bracketed numbers denote the second best result. Non-DS denotes the C3D-Transformer-spectral-multitask-regressor model trained without domain-specific learning strategy while DS denotes the same model trained with domain-specific learning strategy.}  
\label{tb:chalearn}
\end{table*}
\setlength{\tabcolsep}{1.4pt}

\subsubsection{Ablation studies}
\label{subsec:ablation}

\setlength{\tabcolsep}{2pt}
\begin{table*}[t!]
	\begin{center}
\resizebox{1\linewidth}{!} {
		\begin{tabular}{|l| c c c c c c|}
			\toprule
Methods & Extra & Agree & Consc & Neuro & Open & Avg.   \\
\hline \hline

C3D \cite{tran2015learning} & 0.8892 & 0.8918 & 0.8893 & 0.8919 & 0.8915 & 0.8907  \\

TPN \cite{yang2020temporal}  & 0.8935 & 0.8929 & 0.8855 & 0.8991 & 0.8918 & 0.8926  \\

C3D-Transformer & 0.9052 & 0.9129 & 0.9025 & 0.9098 & 0.9070 & 0.9075  \\

			\bottomrule
		\end{tabular}
        }
	\end{center}
	\caption{ACC results achieved by C3D, TPN and the proposed C3D-Transformer.}  
\label{tb:C3D-Transformer}
\end{table*}
\setlength{\tabcolsep}{1.4pt}

\begin{figure*}
  \begin{center}
  \includegraphics[width=16.8cm]{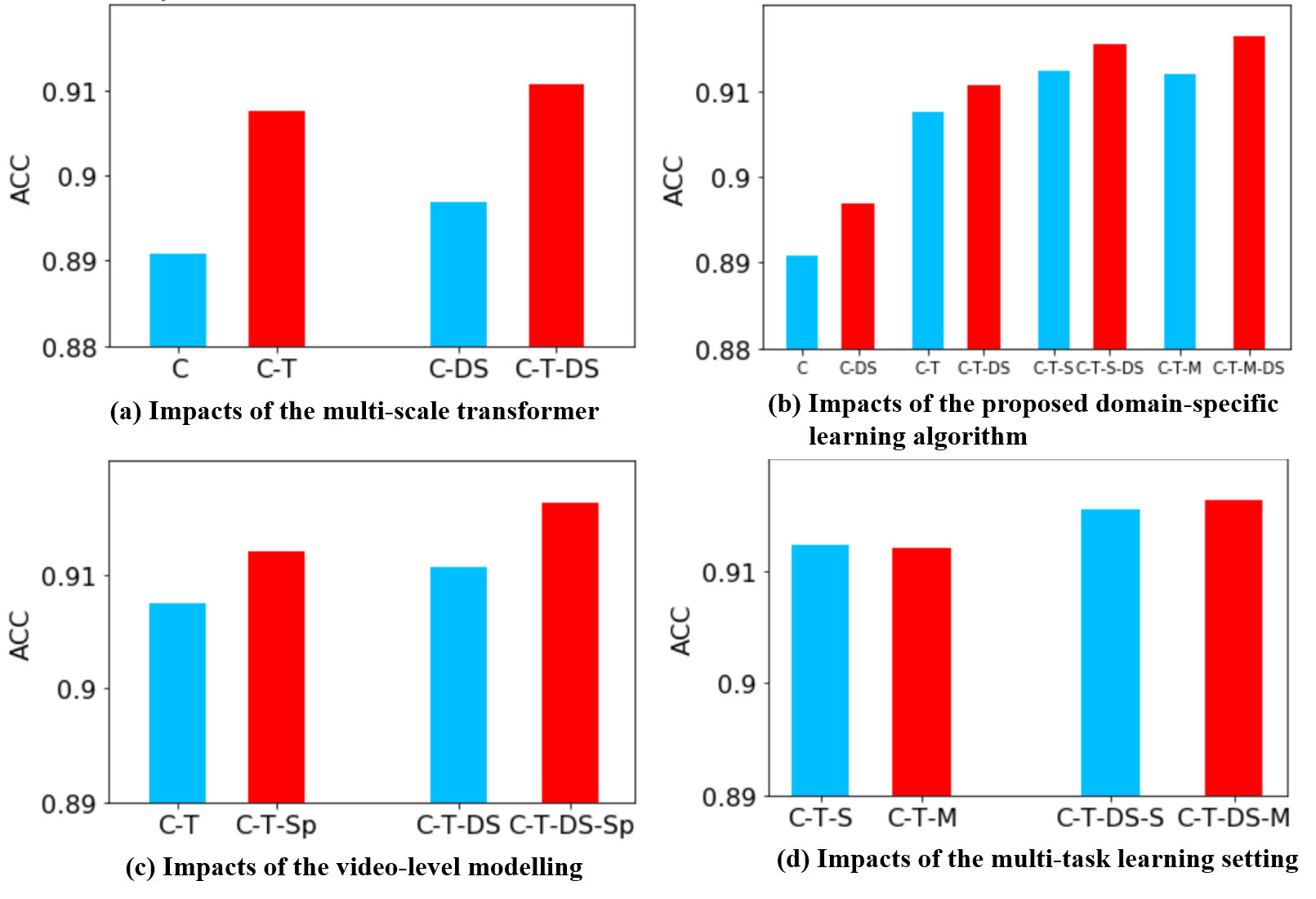}
  \caption{The results of ablation studies.}
  \label{fig:Results}
  \end{center}
\end{figure*}

\noindent We first compare the proposed C3D-Transformer with standard C3D and a standard multi-scale temporal CNNs (Temporal Pyramid Network (TPN) \cite{yang2020temporal}) in Table. \ref{tb:C3D-Transformer} and Fig. \ref{fig:Results}(a). All three systems are trained by standard MSE loss function, and the input of the TPN is set as the same as the C3D-Transformer while we feed $15$ frames for C3D network for each trail. For all systems, the video-level prediction were obtained by averaging the segment-level predictions. It is clear that adding the proposed multi-scale temporal transformer can largely enhance the performance of the C3D network over all five traits, with $1.9\%$ average ACC improvement. Meanwhile, it also outperformed the TPN network for all five traits, with $1.7\%$ average ACC improvement. These results not only demonstrate that personality traits-related clues are contained in multi-scale facial dynamics but also show that the attention operations in transformer can better combine multi-scale cues than standard convolution operations. In short, the proposed C3D-Transformer has strong ability to capture short-term personality-related facial clues.

We then evaluate the impact of the proposed domain-specific learning strategy. As illustrated in Fig. \ref{fig:Results}(b), the ACC results achieved by C3D-Transformer trained with domain-specific learning allows significant improvement for four traits (Extraversion, Conscientiousness, Neuroticism, and Openness), while the achieved similar results as the C3D-Transformer trained with standard MSE for the Extraversion trait. Meanwhile, we also found that the C3D trained with the proposed domain-specific learning strategy outperformed the C3D trained with standard MSE in predicting all five traits. We believe these results show the fact that the domain-specific learning strategy can efficiently remove the noises or personality un-related information from the learned multi-scale dynamic features, and therefore achieved enhanced recognition performances for both C3D and C3D-Transformer models.

As discussed in Sec. \ref{sec:intro}, the hypothesis of this paper is that long-term facial clues are more important for personality traits recognition. Thus, we also compare the results achieved by the C3D-Transformer with and without video-level modelling step. In particular, we extend the spectral encoding to summarize both segment-level features and predictions to generate the video-level prediction. As shown in Fig. \ref{fig:Results}(c), the spectral-based video-level modelling improved the average performance for both C3D-Transformers that are trained with MSE or domain-specific learning strategy. This is because the spectral encoding captures multiple video-level frequencies of the segment-level data, which describe multi-scale facial behaviors for personality traits prediction, which are ignored when simply averaging segment-level predictions. In short, we conclude that a further long-term modelling of the short-term features can improve the video-level personality recognition performance. A similar conclusion also has been made for another task \cite{xu2021two}.

Finally, we evaluate the importance of the multi-task regression by comparing the proposed multi-task learning scheme to the systems that individually predicts each trait, i.e., training five regressors, where each takes the video-level spectral feature for predicting a single trait. According to the results in Fig. \ref{fig:Results}(d), for spectral features of the C3D-Transformer trained with MSE, the proposed multi-task regression generated better performance for Extraversion, Conscientiousness, and Openness traits. Meanwhile, for spectral features of the C3D-Transformer trained with domain-specific learning strategy, the proposed multi-task regression generated better performance for Extraversion, Agreement, Conscientiousness, and Openness traits as well as the average result. Despite the multi-task learning did not provide significant improvements for most conditions, it still enhanced most traits' recognition as well as the average performance, which means this scheme allows the underlying relationship information among five traits, which are personality-related, to be used for personality traits recognition.

\section{Conclusion}

\noindent In this paper, we propose a video-based automatic apparent personality recognition approach which has three main advantages: 1. we propose a novel multi-scale spatio-temporal transformer to learn multi-scale short-term behavioural temporal information of the video; 2. the proposed domain-specific learning strategy can extract the relevant cues from the deep-learned features for each personality trait's recognition; 3. our approach can model the underlying relationship from the video-level representation of the five traits. The experimental results show that although the proposed system avoids complex pre-processing as in \cite{li2020cr} and high computational complexity as in \cite{song2021self}, it still achieved a top performance among existing approaches. Specifically,  we show that: 1. the proposed C3D-Transformer is a superior backbone for learning short-term personality-related facial behaviors from short video segments; 2. the proposed domain-specific learning strategy clearly improved the systems that are learned by standard training process, demonstrating our strategy can enhance the network's ability to learn depression-related cues; 3. the video-level personality modelling provide more reliable predictions than these learned from short-term facial behaviors.

The main limitation of this work is that while the proposed approach can be treated as a general video classification/regression framework, we only evaluated it on personality computing. Also, the short segment-level modelling and video-level modelling were conducted separately rather than trained by an end-to-end scheme. Consequently, our future work would focus on two parts: 1. extending the proposed approach to more video analysis tasks and provide a fair evaluation of it; 2. developing an end-to-end training method allowing our model to jointly model the short-term and long-term facial behaviors for personality recognition.






%


\bibliographystyle{IEEEtran}
\bibliography{IEEEabrv,reference}

\end{document}